# Real time Detection of Lane Markers in Urban Streets

Mohamed Aly
Computational Vision Lab
Electrical Engineering
California Institute of Technology
Pasadena, CA 91125
malaa@caltech.edu

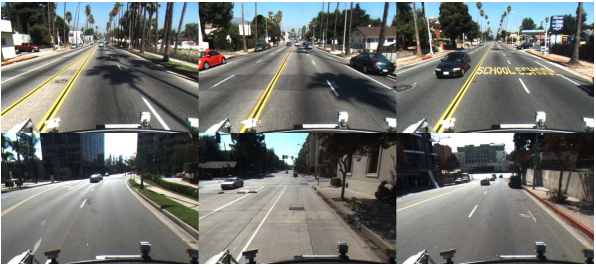

Fig. 1. Challenges of lane detection in urban streets

*Abstract*— We present a robust and real time approach to lane marker detection in urban streets. It is based on generating a top view of the road, filtering using selective oriented Gaussian filters, using RANSAC line fitting to give initial guesses to a new and fast RANSAC algorithm for fitting Bezier Splines, which is then followed by a post-processing step. Our algorithm can detect *all* lanes in still images of the street in various conditions, while operating at a rate of 50 Hz and achieving comparable results to previous techniques.

## I. INTRODUCTION

Car accidents kill about 50,000 people each year in the US. Up to 90% of these accidents are caused by driver faults [1]. Automating driving may help reduce this huge number of human fatalities. One useful technology is lane detection which has received considerable attention since the mid 1980s [15], [8], [13], [10], [2]. Techniques used varied from using monocular [11] to stereo vision [5], [6], using low-level morphological operations [2], [3] to using probabilistic grouping and B-snakes [14], [7], [9]. However, most of these techniques were focused on detection of lane markers on highway roads, which is an easier task compared to lane detection in urban streets. Lane detection in urban streets is especially a hard problem. Challenges include: parked and moving vehicles, bad quality lines, shadows cast from trees, buildings and other vehicles, sharper curves, irregular/strange lane shapes, emerging and merging lanes, sun glare, writings and other markings on the road (e.g. pedestrian crosswalks), different pavement materials, and different slopes (fig. 1).

This paper presents a simple, fast, robust, and effective approach to tackle this problem. It is based on taking a top-view of the image, called the Inverse Perspective Mapping (IPM) [2]. This image is then filtered using selective Gaussian spatial filters that are optimized to detecting vertical lines. This filtered image is then thresholded robustly by keeping only the highest values, straight lines are detected using simplified Hough transform, which is followed by a RANSAC line fitting step, and then a novel RANSAC spline fitting step is performed to refine the detected straight lines and correctly detect curved lanes. Finally, a cleaning and localization step is performed in the input image for the detected splines.

This work provides a number of contributions. First of all, it's robust and real time, running at 50 Hz on 640x480 images on a typical machine with Intel Core2 2.4 GHz machine. . Second, it can detect any number of lane boundaries in the image not just the current lane i.e. it can detect lane boundaries of neighboring lanes as well. This is a first step towards understanding urban road images. Third, we present a new and fast RANSAC algorithm for fitting splines efficiently. Finally, we present a thorough evaluation of our approach by employing hand-labeled dataset of lanes and introducing an automatic way of scoring the detections found by the algorithm. The paper is organized as follows: section II gives a detailed description of the approach. Section III shows the experiments and results, which is followed by a discussion in section IV. Finally, a conclusion is given in section V.

## II. APPROACH

### A. Inverse Perspective Mapping (IPM)

The first step in our system is to take generate a top view of the road image [2]. This has two benefits:

1) We can get rid of the perspective effect in the image, and so lanes that appear to converge at the horizon line are now vertical and parallel. This uses our main assumption that the lanes are parallel (or close to parallel) to the camera.
2) We can focus our attention on only a subregion of the input image, which helps in reducing the run time considerably.

To get the IPM of the input image, we assume a flat road, and use the camera intrinsic (focal length and optical center) and extrinsic (pitch angle, yaw angle, and height above ground) parameters to perform this transformation. We start by defining a world frame $\{F_w\} = \{X_w, Y_w, Z_w\}$ centered at the camera optical center, a camera frame $\{F_c\} = \{X_c, Y_c, Z_c\}$, and an image frame $\{F_i\} = \{u, v\}$ as show in figure 2. We assume that the camera frame $X_c$ axis stays in the world frame $X_w Y_w$ plane i.e. we allow for a pitch angle $\alpha$ and yaw angle $\beta$ for the optical axis but no roll. The height of the camera frame above the ground plane is $h$. Starting from any point in the image plane $^iP = \{u, v, 1, 1\}$, it can be shown that its projection on the road plane can be found by applying the homogeneous transformation $^g_i T =$

$$h \begin{bmatrix} -\frac{1}{f_u}c_2 & \frac{1}{f_v}s_1 s_2 & \frac{1}{f_u}c_u c_2 - \frac{1}{f_v}c_v s_1 s_2 - c_1 s_2 & 0 \\ \frac{1}{f_u}s_2 & \frac{1}{f_v}s_1 c_1 & -\frac{1}{f_u}c_u s_2 - \frac{1}{f_v}c_v s_1 c_1 - c_1 c_2 & 0 \\ 0 & \frac{1}{f_v}c_1 & -\frac{1}{f_v}c_v c_1 + s_1 & 0 \\ 0 & -\frac{1}{hf_v}c_1 & \frac{1}{hf_v}c_v c_1 - \frac{1}{h}s_1 & 0 \end{bmatrix}$$

i.e. $^gP = {^g_i T}\, ^iP$ is the point on the ground plane corresponding to $^iP$ on the image plane, where $\{f_u, f_v\}$ are the horizontal and vertical focal lengths, respectively, $\{c_u, c_v\}$ are the coordinates of the optical center, and $c_1 = \cos \alpha$, $c_2 = \cos \beta$, $s_1 = \sin \alpha$, and $s_2 = \sin \beta$. These transformations can be efficiently calculated in matrix form for hundreds of points. The inverse of the transform can be easily found to be $^i_g T =$

$$\begin{bmatrix} f_u c_2 + c_u c_1 s_2 & c_u c_1 c_2 - s_2 f_u & -c_u s_1 & 0 \\ s_2(c_v c_1 - f_v s_1) & c_2(c_v c_1 - f_v s_1) & -f_v c_1 - c_v s_1 & 0 \\ c_1 s_2 & c_1 c_2 & -s_1 & 0 \\ c_1 s_2 & c_1 c_2 & -s_1 & 0 \end{bmatrix}$$

where again starting from a point on the ground $^gP = \{x_g, y_g, -h, 1\}$ we can get its subpixel coordinates on the image frame by $^iP = {^i_g T}\, ^gP$ and then rescale the homogeneous part. Using these two transformations, we can project a window of interest from the input image onto the ground plane. Figure 3 shows a sample IPM image. The left side shows the original image (640x480 pixels), with the region of interest in red, and the right image shows the transformed IPM image (160x120 pixels). As shown, lanes in the IPM image have fixed width in the image and appear as vertical, parallel straight lines.

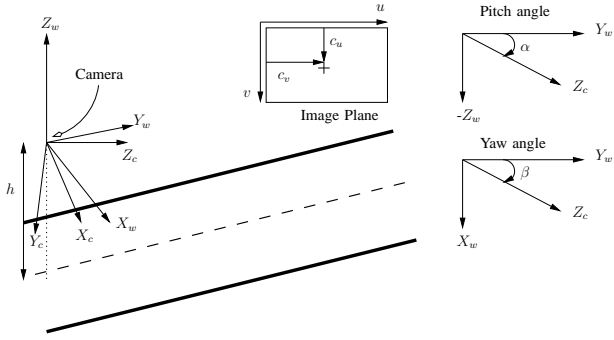

Fig. 2. IPM coordinates. Left: the coordinate axes (world, camera, and image frames). Right: definition of pitch $\alpha$ and yaw $\beta$ angles.

### B. Filtering and Thresholding

The transformed IPM image is then filtered by a two dimensional Gaussian kernel. The vertical direction is a smoothing Gaussian, whose $\sigma_y$ is adjusted according to the required height of lane segment (set to the equivalent of 1m in the IPM image) to be detected: $f_v(y) = \exp(-\frac{1}{2\sigma_y^2} y^2)$. The horizontal direction is a second-derivative of Gaussian, whose $\sigma_x$ is adjusted according to the expected width of the lanes (set to the equivalent of 3 inches in the IPM image): $f_u(x) = \frac{1}{\sigma_x^2} \exp(-\frac{x^2}{2\sigma_x^2})(1 - \frac{x^2}{\sigma_x^2})$. The filter is tuned specifically for vertical bright lines on dark background of specific width, which is our assumption of lanes in the IPM image, but can also handle quasi-vertical lines which produce considerable output after the thresholding process.

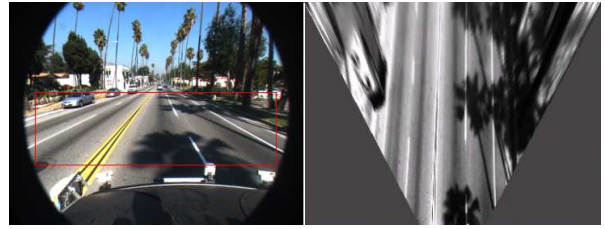

Fig. 3. IPM sample. Left: input image with region of interest in red. Right: the IPM view.

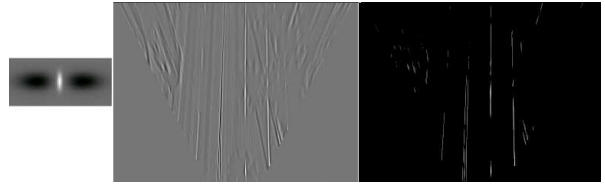

Fig. 4. Image filtering and thresholding. Left: the kernel used for filtering. Middle: the image after filtering. Right: the image after thresholding

Using this separable kernel allows for efficient implementation, and is much faster than using a non-separable kernel. Figure 4 shows the resulting 2D kernel used (left) and the resulting filtered image (middle). As can be seen from the filtered image, it has high response to lane markers, and so we only retain the highest values. This is done by selecting the $q\%$ quantile value from the filtered image, and removing all values below this threshold i.e. we only keep the highest $(q-1)\%$ of the values. In our experiments, $q$ is set to 97.5% in the experiments. The thresholded image is not binarized i.e. we keep the actual pixel values of the thresholded image, which is the input to the following steps. In this step, we use the assumption that the vehicle is parallel/near parallel to the lanes. Figure 4 (right) shows the result after thresholding.

### C. Line Detection

This stage is concerned with detecting lines in the thresholded image. We use two techniques: a simplified version of Hough Transform to count how many lines there are in the image, followed by a RANSAC [4] line fitting to robustly fit these lines. The simplified Hough transform gets a sum of values in each column of the thresholded filtered image. This sum is then smoothed by a Gaussian filter, local maxima are detected to get positions of lines, and then this is further refined to get sub-pixel accuracy by fitting a parabola to the local maxima and its two neighbors. At last, nearby lines are grouped together to eliminate multiple responses to the same line. Figure 5 shows the result of this step.

The next step is getting a better fit for these lines using RANSAC line fitting. For each of the vertical lines detected above, we focus on a window around it (white box in left of fig. 6), and run the RANSAC line fitting on that window.

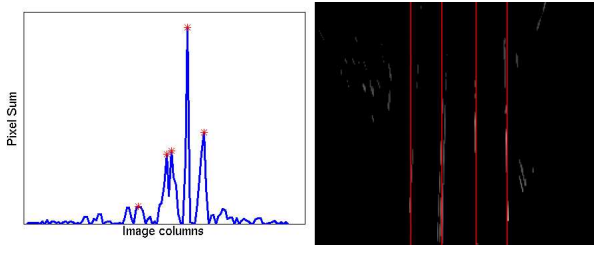

Fig. 5. Hough line grouping. Left: the sum of pixels for each column of the thresholded image with local maxima in red. Right: detected lines after grouping.

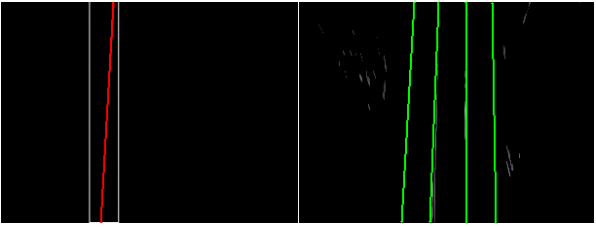

Fig. 6. RANSAC line fitting. Left: one of four windows (white) around the vertical lines from the previous step, and the detected line (red). Right: the resulting lines from the RANSAC line fitting step.

Figure 6 (right) shows the result of RANSAC line fitting on the sample image.

### D. RANSAC Spline Fitting

The previous step gives us candidate lines in the image, which are then refined by this step. For each such line, we take a window around it in the image, on which we will be running the spline fitting algorithm. We initialize the spline fitting algorithm with the lines from the previous step, which is a good initial guess for this step, if the lanes are straight. The spline used in these experiments is a third degree Bezier spline [12], which has the useful property that the control points form a bounding polygon around the spline itself.

The third degree Bezier spline is defined by:

$$\begin{aligned} Q(t) &= T(t)MP \\ &= \begin{bmatrix} t^3 & t^2 & t & 1 \end{bmatrix} \begin{bmatrix} -1 & 3 & -3 & 1 \\ 3 & -6 & 3 & 0 \\ -3 & 3 & 0 & 0 \\ 1 & 0 & 0 & 0 \end{bmatrix} \begin{bmatrix} P_0 \\ P_1 \\ P_2 \\ P_3 \end{bmatrix} \end{aligned}$$

where $t \in [0,1]$, $Q(0) = P_0$ and $Q(1) = P_3$ and the points $P_1$ and $P_2$ control the shape of the spline (figure 7).

Algorithm 1 describes the RANSAC spline fitting algorithm. The basic three function inside the main loop are:

1) getRandomSample(): This function samples from the points available in the region of interest passed to the RANSAC step. We use a weighted sampling approach, with weights proportional to the pixel values of the thresholded image. This helps in picking more the relevant points i.e. points with higher chance of belonging to the lane.

2) fitSpline(): This takes a number of points, and fits a Bezier spline using a least squares method. Given a

**Algorithm 1** RANSAC Spline Fitting
**for** $i = 1$ to $numIterations$ **do**
  $points$=getRandomSample()
  $spline$=fitSpline($points$)
  $score$=computeSplineScore($spline$)
  **if** $score > bestScore$ **then**
    $bestSpline = spline$
  **end if**
**end for**

sample of $n$ points, we assign a value $t_i \in [0,1]$ to each point $p_i = (u_i, v_i)$ in the sample, where $t_i$ is proportional to cumulative sum of the euclidean distances from point $p_i$ to the first point $p_1$. Define a point $p_0 = p_1$, we have:

$$t_i = \frac{\sum_{j=1}^{i} d(p_j, p_{j-1})}{\sum_{j=1}^{n} d(p_j, p_{j-1})} \text{ for } t_i = 1..n$$

where $d(p_i, p_j) = \sqrt{(u_i - u_j)^2 + (v_i - v_j)^2}$. This forces $t_1 = 0$ and $t_n = 1$ which corresponds to the first and last point of the spline, respectively. Next, we define the following matrices:

$$Q = \begin{bmatrix} p_1 \\ ... \\ p_n \end{bmatrix}$$

$$T = \begin{bmatrix} t_1^3 & t_1^2 & t_1 & 1 \\ & ... & & \\ t_n^3 & t_n^2 & t_n & 1 \end{bmatrix}$$

and solve for the matrix $P$ using the pseudo-inverse:

$$P = (TM)^{\dagger} Q$$

This gives us the control points for the spline that minimizes the sum of squared error of fitting the sampled points.

3) computeSplineScore(): In normal RANSAC, we would be interested in computing the normal distance from every point to the third degree spline to decide the goodness of that spline, however this would require solving a fifth degree equation for every such point. Instead, we decided to follow a more efficient approach. It computes the score (measure of goodness) of the spline by rasterizing it using an efficient iterative way [12], and then counting the values of pixels belonging to the spline. It also takes into account the straightness and length of the spline, by penalizing shorter and more curved splines. Specifically, the score is computed as:

$$score = s(1 + k_1 l' + k_2 \theta')$$

where $s$ is the raw score for the spline (the sum of pixel values of the spline), $l$' is the normalized spline length measure defined as $l' = (l/v) - 1$ where $l$ is the spline length and $v$ is the image height and so $l' = 0$ means we have a longer spline and $l' = -1$ means a shorter spline, $\theta'$ is the normalized spline "curveness" measure defined by $\theta' = (\theta - 1)/2$ whereas $\theta$ is the mean of the cosine of angles between lines joining the spline's control point i.e. $\theta = (\cos\theta_1 + \cos\theta_2)/2$, and $k_1$ and $k_2$ are regularization factors, see figure 7. This scoring formula makes sure we favor longer and straighter splines than shorter and curvier ones, where longer and straighter splines are penalized less than shorted curvier ones.

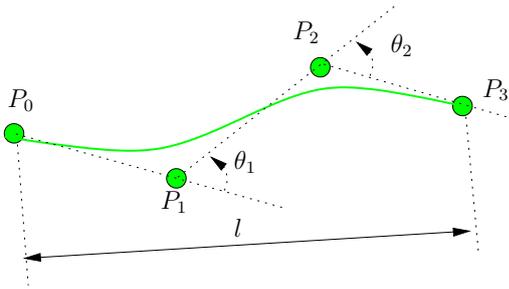

Fig. 7. Spline score computation.

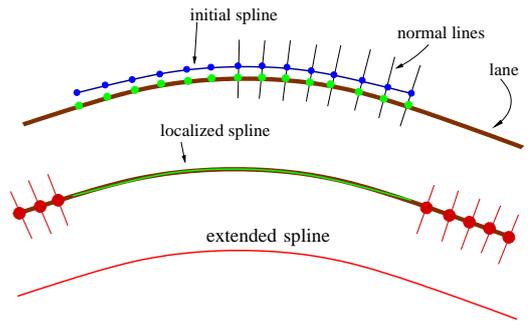

Fig. 9. Spline localization and extension.

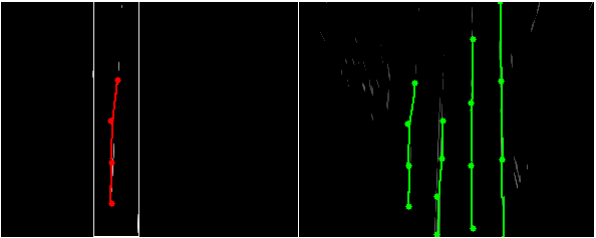

Fig. 8. RANSAC Spline fitting. Left: one of four windows of interest (white) obtained from previous step with detected spline (red). Right: the resulting splines (green) from this step

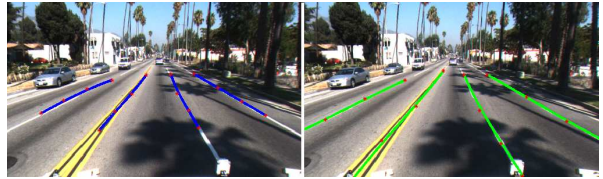

Fig. 10. Post-processing splines. Left: splines before post-processing in blue. Right: splines after post-processing in green. They appear longer and localized on the lanes.

Figure 8 shows a sample result for the algorithm. The left side shows a window of interest (white) around the lines output from the RANSAC line step, with the detected spline in red. The right side shows the four output splines in green.

### E. Post-processing

The final step of the algorithm is to post-process the output of the previous stage to try to better localize the spline and extend it in the image, figure 9. This is done both in the IPM image, and in the original image after back projecting the splines from the IPM space to the image space. Here we perform three steps:

*1) Localization:* We start with the initial spline (blue spline in figure 9), and then we sample points on the spline (blue points), extend a line segment through these sampled points that are normal to the spline tangent direction at that point (black line segments). Then, we get the grayscale profile for this line segment by computing the pixel locations that this line passes through, convolve that with a smoothing Gaussian kernel, and look for local maxima of the result. This should give us better localization for points on the spline to give better fit for the road lanes (green points). In addition, one more check is performed on the angle change of the newly detected point, and this new point is rejected if it lies so far from the expected location. Finally, we refit the spline with the localized points (green spline).

*2) Extension:* After the spline's position has been improved, we perform an extension in the IPM and original images, in order to give an even better fit of the lane. This is done similarly by looking both forward and backward from the spline end points along the tangent direction (red points), and creating line segments through the normal direction (red line segments), and finding the peak of convolving the grayscale profile of these segments with the smoothing Gaussian filter. The new peak is not accepted if it's below a certain threshold (homogeneous area with no lines in it), or if the orientation change from the dominant spline orientation exceeds a certain threshold, in which case the extension process stops.

*3) Geometric Checks:* After each of the previous two steps, we also perform geometrical checks on the localized and extended splines, to make sure they are not very curved or very short, in which case they are replaced by the corresponding line from the RANSAC line fitting stage. Checks are also made to make sure fitted splines are near vertical in the IPM image, otherwise they are rejected as valid splines. Figure 10 shows the results before and after post-processing the splines.

## III. EXPERIMENTS

### A. Setup

We collected a number of clips on different types of urban streets, with/without shadows, and on straight and curved streets. Unlike previous papers that would just mention rough percentages of detection rates, and in order to get an accurate quantitative assessment of the algorithm, we hand-labeled all visible lanes in four of these clips, totaling 1224 labeled frames containing 4172 marked lanes (table I). The system was prototyped using Matlab, and implemented in C++ using the open source OpenCV library. These clips are quite challenging, for clip #1 has a lot of curvatures and some writings on the street, clip #2 has different pavement types and the sun is facing the vehicle, clip #3 has lots of shadows (at the beginning) and passing cars, and finally clip #4 has street writings and passing vehicles as well (fig. 1).

TABLE I
DATASETS

| Clip# | name | #frames | #lane boundaries |
|---|---|---|---|
| 1 | cordova1 | 250 | 919 |
| 2 | cordova2 | 406 | 1048 |
| 3 | washington1 | 336 | 1274 |
| 4 | washington2 | 232 | 931 |
| Total | | 1224 | 4172 |

TABLE II
RESULTS FOR 2-LANES MODE

| Clip | #total | #detected | correct rate | false pos. rate | fp/frame |
|---|---|---|---|---|---|
| 1 | 466 | 467 | 97.21% | 3.00% | 0.056 |
| 2 | 472 | 631 | 96.16% | 38.38% | 0.443 |
| 3 | 639 | 645 | 96.70% | 4.72% | 0.089 |
| 4 | 452 | 440 | 95.13% | 2.21% | 0.043 |
| Total | 2026 | 2183 | **96.34%** | 11.57% | 0.191 |

TABLE III
RESULTS FOR ALL-LANES

| Clip | #total | detected | correct rate | false pos. rate | fp/frame |
|---|---|---|---|---|---|
| 1 | 919 | 842 | 91.62% | 5.66% | 0.208 |
| 2 | 1048 | 1322 | 85.50% | 40.64% | 1.049 |
| 3 | 1274 | 1349 | 92.78% | 13.11% | 0.497 |
| 4 | 931 | 952 | 93.66% | 8.59% | 0.345 |
| Total | 4172 | 4517 | **90.89%** | 17.38% | 0.592 |

The detection results shown in the next section are computed automatically using the hand-labeled data. In each frame, each detected lane boundary is compared to ground truth lanes, and a check is made to decide if it is a correct or false detection. To check if two splines $s_1$ and $s_2$ are the same i.e. represent the same lane boundary, we sample points on both of them $p_i^1$ and $p_i^2$. For every point $p_i^1$ on the first spline, we compute the nearest point $p_j^2$ on the second spline and compute the distance $d_i^1$ between the two points. We do the same for the second spline, where for every such point we get the nearest distance $d_i^2$ to the first spline. We then compute the median distances $\hat{d}_1$ & $\hat{d}_2$ and the mean distances $\bar{d}_1$ & $\bar{d}_2$. Now to decide whether they are the same, we require that both

$$\min(\hat{d}_1, \hat{d}_2) \leq t_1$$

&

$$\min(\bar{d}_1, \bar{d}_2) \leq t_2$$

be satisfied. In our experiments, we used $t_1 = 20$ and $t_2 = 15$.

*B. Results*

We ran the algorithm in two different modes: **1) 2-lanes mode**: detecting only the two lane boundaries of the current lane, which is similar to previous approaches; and **2) all-lanes mode**: detecting all visible lanes in the image. In the first mode, we just focus on the middle of the IPM image by clipping the left and right parts, while in the second mode we work on the whole IPM image. Tables II and III show results for the two modes. The fist column shows the total number of lane boundaries in each clip, the second shows the number of detected lane boundaries, the third the correct detection rate, followed by the false positive rate, and finally the false positive/frame rate. Figure 11 shows some detection results samples from these clips. The complete videos can be accessed online at http://www.vision.caltech.edu/malaa/research/iv08.

IV. DISCUSSION

The results show the effectiveness of our algorithm in detecting lanes on urban streets with varying conditions. Our algorithm doesn't use tracking yet i.e. these are the results of detecting lanes in each image independently without utilizing any temporal information. However, when detecting only the lane boundaries of the current lane, we achieve comparable results to other algorithms (e.g. [**?**], [7]), which used both detection and tracking. We also achieve good results for detecting all the visible lane boundaries, which is a first step towards urban road understanding, and which was not attempted before (as far as we know). We get excellent results in clear conditions, however we get some false positives due to stop lines at cross streets, at cross walks, near passing cars, see figure 12.

False positives are mostly found when driving on the right lane of the street with no right lane boundary, and we detect the curb as the right lane boundary (fig. 12), and that's the reason for the high false positive rate in clip #2. However, this is not a real problem, as the curb is really a lane boundary, but not painted as such, and this won't affect the objectives of the algorithm to detect lane boundaries.

In the current algorithm, we only work on the red channel, which gives us better images for white and yellow lanes than converting it to grayscale. However, there is plenty to be done to further improve this algorithm. We plan on using the color information to classify different lane boundaries: white solid lines, double yellow lines, ..etc. This will also allow us to remove the false positives due to curbs being detected as lanes, as well as confusing writings on the streets, which are usually in yellow and can be filtered out. Furthermore, we plan on employing tracking on top of the detection step, which will help get rid of a lot of these false positives.

V. CONCLUSION

We proposed an efficient, real time, and robust algorithm for detecting lanes in urban streets. The algorithm is based on taking a top view of the road image, filtering with Gaussian kernels, and then using line detection and a new RANSAC spline fitting technique to detect lanes in the street, which is followed by a post-processing step. Our algorithm can detect *all* lanes in still images of urban streets and works at high rates of 50 Hz. We achieved comparable results to other algorithms that only worked on detecting the current lane boundaries, and good results for detecting all lane boundaries.

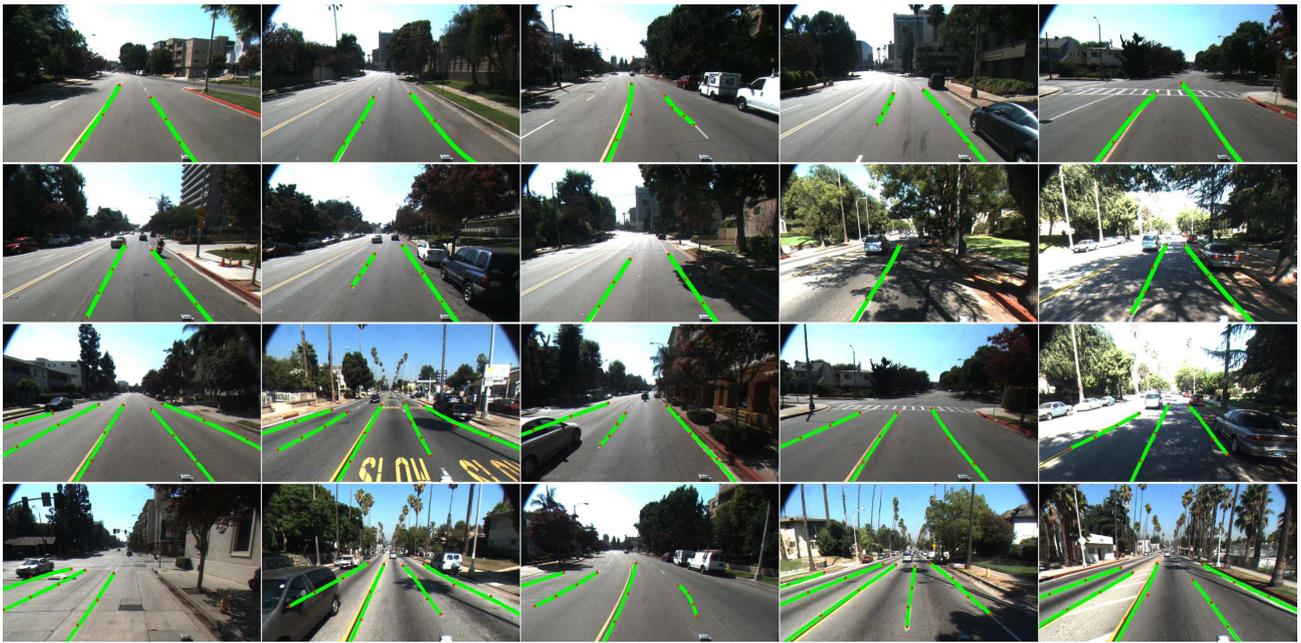

Fig. 11. Detection samples show robustness in presence of shadows, vehicles, curves, different road structures and pavements. First two rows are for the 2-lanes mode, and last two rows are for the all-lanes mode

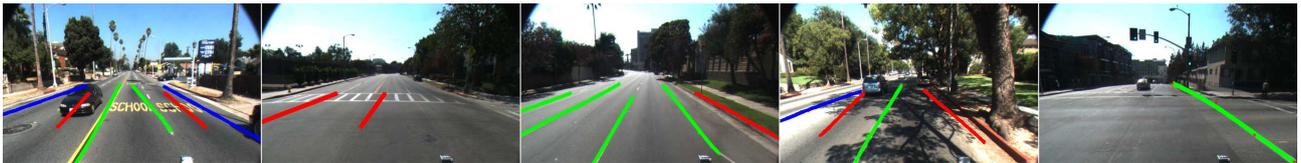

Fig. 12. False detections samples, showing confusion due to street writings, crosswalks, right curb, vehicles, and stop lines on cross streets. Correct detections are in green, false positives are in red, while false negatives (missed lanes) are in blue.

REFERENCES

[1] Car accidents. http://en.wikipedia.org/wiki/car_accident.
[2] M. Bertozzi and A. Broggi. Real-time lane and obstacle detection on the gold system. In *Intelligent Vehicles Symposium, Proceedings of the IEEE*, pages 213–218, 19-20 Sept. 1996.
[3] M. Bertozzi, A. Broggi, G. Conte, and A. Fascioli. Obstacle and lane detection on argo. In *Intelligent Transportation System, IEEE Conference on*, pages 1010–1015, 9-12 Nov. 1997.
[4] David A. Forsyth and Jean Ponce. *Computer Vision: A modern approach*. Prentice Hall, 2002.
[5] U. Franke, D. Gavrila, S. Gorzig, F. Lindner, F. Puetzold, and C. Wohler. Autonomous driving goes downtown. *Intelligent Systems and Their Applications, IEEE*, 13(6):40–48, Nov.-Dec. 1998.
[6] U. Franke and I. Kutzbach. Fast stereo based object detection for stop & go traffic. In *Intelligent Vehicles Symposium, Proceedings of the IEEE*, pages 339–344, 19-20 Sept. 1996.
[7] Zu Kim. Realtime lane tracking of curved local road. In *Intelligent Transportation Systems, Proceedings of the IEEE*, pages 1149–1155, Sept. 17-20, 2006.
[8] K. Kluge. Extracting road curvature and orientation from image edge points without perceptual grouping into features. In *Intelligent Vehicles Symposium, Proceedings of the*, pages 109–114, 24-26 Oct. 1994.
[9] C. Kreucher and S. Lakshmanan. Lana: a lane extraction algorithm that uses frequency domain features. *Robotics and Automation, IEEE Transactions on*, 15(2):343–350, April 1999.
[10] J.C. McCall and M.M. Trivedi. Video-based lane estimation and tracking for driver assistance: survey, system, and evaluation. *Intelligent Transportation Systems, IEEE Transactions on*, 7(1):20–37, March 2006.
[11] D. Pomerleau. Ralph: rapidly adapting lateral position handler. *IEEE Symposium on Intelligent Vehicles*, pages 506–511, 25-26 Sep 1995.
[12] David Solomon. *Curves and Surfaces for Computer Graphics*. Springer, 2006.
[13] C. Taylor, J. seck, R. Blasi, and J. Malik. A comparative study of vision-based lateral control strategies for autonomous highway driving. *International Journal of Robotics Research*, 1999.
[14] Hong Wang and Qiang Chen. Real-time lane detection in various conditions and night cases. In *Intelligent Transportation Systems, Proceedings of the IEEE*, pages 1226–1231, Sept. 17-20, 2006.
[15] A. Zapp and E. Dickmanns. A curvature-based scheme for improved road vehicle guidance by computer vision. In *Proc. SPIE Conference on Mobile Robots*, 1986.